\documentclass{article}
\pdfoutput=1
\usepackage{graphicx}
\usepackage{hyperref}
\usepackage{multirow}
\usepackage[hyperref=true,style=authoryear,backend=bibtex]{biblatex}
\usepackage{url}
\usepackage[table,xcdraw]{xcolor}
\usepackage{enumerate}
\usepackage{amsmath}
\usepackage{amsfonts}
\usepackage{mdframed}

\begin{document}

\title{
  Making Sense of Hidden Layer Information in Deep Networks by Learning
  Hierarchical Targets
}
\author{
  Abhinav Tushar \\
  Department of Electrical Engineering \\
  Indian Institute of Technology, Roorkee \\
  \texttt{abhinav.tushar.vs@gmail.com}
}
\date{}
\maketitle

\begin{abstract}
  This paper proposes an architecture for deep neural networks with hidden layer
  branches that learn targets of lower hierarchy than final layer targets. The
  branches provide a channel for enforcing useful information in hidden layer
  which helps in attaining better accuracy, both for the final layer and hidden
  layers. The shared layers modify their weights using the gradients of all cost
  functions higher than the branching layer. This model provides a flexible
  inference system with many levels of targets which is modular and can be used
  efficiently in situations requiring different levels of results according to
  complexity. This paper applies the idea to a text classification task on
  \emph{20 Newsgroups} data set with two level of hierarchical targets and a
  comparison is made with training without the use of hidden layer branches.
\end{abstract}

\vspace{0.2in}

\begin{mdframed}[backgroundcolor=gray!10]
  {\bf Author's Note}

  \noindent
  {\sl September, 2016}\\

  \noindent
  This document essentially was (May 2015) a hasty write up of a project for a
  course on artificial neural networks during my undergraduate studies. I am
  adding this note here to point out mistakes which are detrimental to writings.
  I have kept the original content intact, adding only this box.

  Firstly, the document doesn't really use the terms ({\sl information}, {\sl
    deep networks}) from the title well in the analysis. Talking about the idea
  itself, there is a similar concept of auxiliary classifier in literature which
  uses [same] targets at lower levels to improve performance (See
  \href{https://arxiv.org/pdf/1409.4842.pdf}{arXiv:1409.4842v1 [cs.CV]}
  for example). -1 to literature review. Furthermore, the comparison is not
  rigorous enough to back up the claims and needs more meaningful test.
\end{mdframed}

\section{Introduction}
Deep neural networks aim at learning multiple level of features by using larger
number of hidden layers as compared to shallow networks. Using many layers,
higher order features can be automatically learned without the need of any
domain specific feature engineering. This makes them more generalized inference
systems. They are effective at learning features from raw data which would have
required much efforts to pre process in case of shallow networks, for example, a
recent work \parencite{Zhang2015} demonstrated deep temporal convolutional
networks to learn abstract text concepts from character level inputs.

However, having multiple layers, deep networks are not easy to train. Few of the
problems are, getting stuck in local optima, problem of vanishing gradients etc.
If the hyperparameters of networks are not engineered properly, deep networks
also tend to overfit. The choice of activation functions \parencite{Glorot2010}
as well as proper initialization of weights \parencite{Sutskever2013} plays
important role in the performance of deep networks.

Several methods have been proposed to improve the performance of deep networks.
Layer by layer training of Deep Belief Networks \parencite{Hinton2006} uses
unsupervised pre-training of the component Restricted Boltzmann Machines (RBMs)
and further supervised fine tuning of the whole network. Similar models have
been presented \parencite{Bengio2007, Ranzato2007} that pre-train the network
layer by layer and then fine tune using supervised techniques.

Unsupervised pre-training is shown to effectively works as a
regularizer \parencite{Erhan2009, Erhan2010} and increase the performance as
compared to network with randomly initialized weights.

This paper explores the idea of training a deep network to learn hierarchical
targets in which lower level targets are learned from taps in lower hidden
layers, while the highest level of target (which has highest details) is kept at
the final layer of the network. The hypothesis is that this architecture should
learn meaningful representation in hidden layers too, because of the branchings.
This can be helpful since the same model can be used as an efficient inference
system for any level of target, depending on the requirement. Also, the
\emph{meaningful} information content of hidden layer activations can be helpful
in improving the overall performance of the network.

The following section presents the proposed deep network with hidden layer
branchings. Section~\ref{sec:results} provides the experimental results on
\emph{20 Newsgroups} data set \footnote{The dataset can be downloaded here
  \url{qwone.com/~jason/20Newsgroups/}} along with the details of the network
used in the experiment. Section~\ref{sec:conclusion} contains the concluding
remarks and scope of future work is given in Section~\ref{sec:future}.

\section{Proposed Network Architecture}
\label{sec:tapped}

\begin{figure*}[h]
  \centering
  \includegraphics[width=0.8\textwidth]{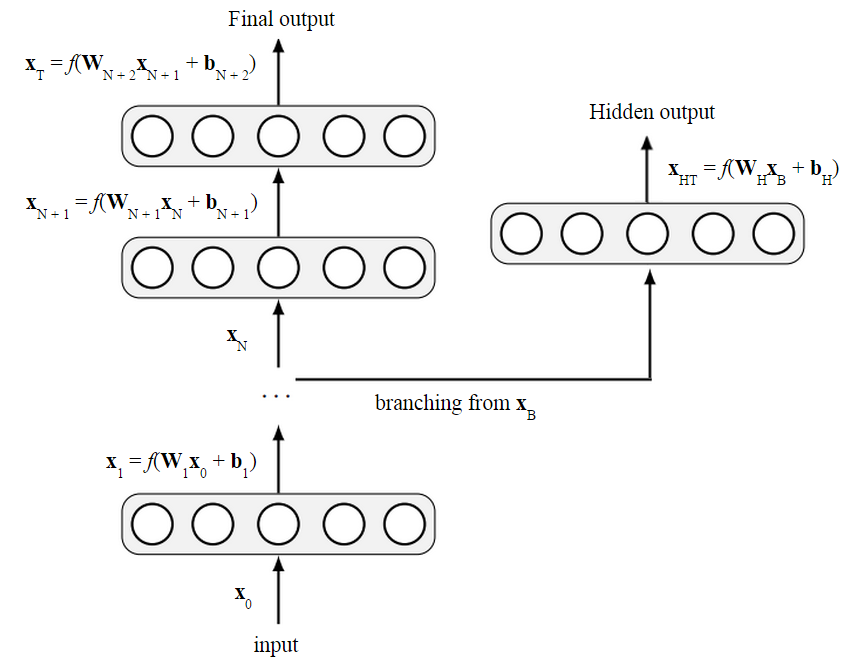}
  \caption{Branched Network Structure}
  \label{fig:branchednet}
\end{figure*}

In the proposed network, apart from the final target layer, one (or more) target
layer are branched from the hidden layers. A simple structure with one branching
is shown in Figure~\ref{fig:branchednet}. The target layers are arranged in a
hierarchical fashion with the most detailed targets being farthest form the
input, while trivial targets closer to the input layer. The network will learn
both the final layer outputs as well as hidden layer outputs. The following sub
section explains the learning algorithm using the example network in
Figure~\ref{fig:branchednet}.

\subsection{Learning Algorithm}
\label{sec:algo}

The network learns using Stochastic Gradient Descent. There are two costs to
minimize, the first being that of final target and second of hidden target. For
the network shown in the Figure~\ref{fig:branchednet}, the network has a branch
from the layer whose output is $\boldsymbol{x}_{B}$. Weights and biases from
$\boldsymbol{W}_{B + 1}, \boldsymbol{b}_{B + 1}$ to $\boldsymbol{W}_{N + 2},
\boldsymbol{b}_{N + 2}$ are updated using the final target layer cost function
only, while $\boldsymbol{W}_{H}$ and $\boldsymbol{b}_{H}$ are updated using only
the hidden layer cost function.

\begin{gather}
  \boldsymbol{W}_{i} \leftarrow \boldsymbol{W}_{i} - \eta \frac{\partial
    \boldsymbol{C}}{\partial \boldsymbol{W}_{i}} \\
  \boldsymbol{b}_{i} \leftarrow \boldsymbol{b}_{i} - \eta \frac{\partial
    \boldsymbol{C}}{\partial \boldsymbol{b}_{i}}
\end{gather}

Here, $\boldsymbol{C}$ is the hidden or final target cost function, depending on
which weights are to be minimized. For the weights that are shared for both
targets, i.e. weights and biases from $\boldsymbol{W}_{1}, \boldsymbol{b}_{1}$
to $\boldsymbol{W}_{B}, \boldsymbol{b}_{B}$, the training uses both cost
function and an averaged update is done for these parameters. If final target
cost is $\boldsymbol{C}_{F}$ and hidden target cost is $\boldsymbol{C}_{H}$,
then the updates are:

\begin{gather}
  \boldsymbol{W}_{i} \leftarrow \boldsymbol{W}_{i} - \eta \Big(\alpha
  \frac{\partial \boldsymbol{C}_F}{\partial \boldsymbol{W}_{i}} + (1 - \alpha)
  \frac{\partial \boldsymbol{C}_H}{\partial \boldsymbol{W}_{i}}\Big) \\
  \boldsymbol{b}_{i} \leftarrow \boldsymbol{b}_{i} - \eta \Big(\alpha
  \frac{\partial \boldsymbol{C}_F}{\partial \boldsymbol{b}_{i}} + (1 - \alpha)
  \frac{\partial \boldsymbol{C}_H}{\partial \boldsymbol{b}_{i}}\Big)
\end{gather}

A value of $\alpha = 0.5$ gives equal weights to both gradients. This value will
be used in the experiment in this paper.

\subsection{Features of the network}

\begin{itemize}

\item \textbf{Performance}
  Representation of meaningful data in hidden layers governed by the hidden
  layer branchings helps by providing features for higher layers and thus
  improves the overall performance of the network.

\item \textbf{Hierarchical targets}
  Different target branches, arranged in hierarchy of details, help in problems
  demanding scalability in level of details of targets.

\item \textbf{Modularity}
  The hidden layer targets lead to storage of meaningful content in hidden
  layers and thus, the network can be separated (recombined) from (with) the
  branch joints without loss of the learned knowledge.

\end{itemize}

\section{Experimental Results}
\label{sec:results}

Hidden layer taps can be exploited only if the problem has multiple and
hierarchical targets. It can also work when it is possible to \emph{degrade} the
resolution (or any other parameter related to details) of output to create
hidden layer outputs. This section explores the performance of the proposed
model on \emph{20 Newsgroups} dataset.

\subsection{Data set}

The data set has newsgroup posts from 20 newsgroups, thus resulting in a 20
class classification problem. According to the newsgroup topics, the 20 classes
were partitioned in 5 primitive classes (details are in
Table~\ref{tbl:classes}). The final layer of the network is made to learn the 20
class targets, while the hidden layer branching is made to learn the cruder, 5
class targets. The dataset has \textbf{18846} instances. Out of these,
\textbf{14314} were selected for training, while the other \textbf{4532}
instances were kept for testing.

\begin{table}[h]
  \begin{center}
    \begin{tabular}{l c r}
      \hline
      \textbf{Primitive class} & \textbf{Final class} & \textbf{Newsgroup topic} \\
      \hline
      \multirow{5}{*}{1} & 1 & comp.graphics \\
      & 2 & comp.os.ms-windows.misc \\
      & 3 & comp.sys.ibm.pc.hardware \\
      & 4 & comp.sys.mac.hardware \\
      & 5 & comp.windows.x \\
      \hline
      \multirow{4}{*}{2} & 6 & rec.autos \\
      & 7 & rec.motorcycles \\
      & 8 & rec.sport.baseball \\
      & 9 & rec.sport.hockey \\
      \hline
      \multirow{4}{*}{3} & 10 & sci.crypt \\
      & 11 & sci.electronics \\
      & 12 & sci.med \\
      & 13 & sci.space \\
      \hline
      \multirow{4}{*}{4} & 14 & talk.politics.guns \\
      & 15 & talk.politics.mideast \\
      & 16 & talk.politics.misc \\
      & 17 & talk.religion.misc \\
      \hline
      \multirow{3}{*}{5} & 18 & alt.atheism \\
      & 19 & misc.forsale \\
      & 20 & soc.religion.christian \\
      \hline
    \end{tabular}
  \end{center}
  \caption{Classes in data set. \emph{Primitive} classes are used for training
    hidden layer branches, while \emph{Final} classes are used for training
    final layer}
  \label{tbl:classes}
\end{table}

\subsection{Word2Vec preprocessing}

For representing text, a simple and popular model can be made using Bag of Words
(BoW). In this, a vocabulary of words is built from the corpus, and each
paragraph (or instance) is represented by a histogram of frequency of occurrence
of words from the vocabulary. Although being intuitive and simple, this
representation has a major disadvantage while working with neural networks. The
vocabulary length is usually very large, of the order of tens of thousands,
while each chunk of text in consideration has only few of the possible words,
which results in a very sparse representation. Such sparse input representation
can lead to poor learning and high inefficiency in neural networks. A new tool,
Word2Vec \footnote{Python adaptation here
  \url{https://radimrehurek.com/gensim/models/word2vec.html} \parencite{gensim}}
is used to represent words as dense vectors.

Word2Vec is a tool for computing continuous distributed representation of words.
It uses Continuous Bag of Words and Skip-gram methods to learn vector
representations of words using a corpus \parencite{Mikolov20131, Mikolov20132}.
The representations provided by Word2Vec group similar words closer in latent
space. These vectors have properties like \parencite{Mikolov20133}:

\begin{equation*}
  v('king') - v('man') + v('woman') \approx v('queen')
\end{equation*}

Here, $v('word')$ represents the vector of \emph{``word''}. For the problem in
hand, a Word2Vec model with 1000 dimensional vector output was trained using the
entire dataset (removing English language stop words). For making a vector for
representing each newsgroup post, all the words' vectors in the post were
averaged.

\subsection{Network Architecture}

The network used had 4 hidden layers. The number of neurons in the layers were:

\begin{align*}
  \textbf{1000} (input) \Rightarrow 300 &\Rightarrow 200 \Rightarrow 200 \Rightarrow 130 \Rightarrow \textbf{20} (target)\\
  &\Rightarrow \textbf{5} (hidden target)
  \end{align*}

From hidden layer 1 (with 300 neurons), a branch was created to learn hidden
target. The weights and biases are:

$\boldsymbol{W_{N}, b_{N}}$ for connections from layer $N-1$ to layer $N$.

$\boldsymbol{W_{H}, b_{H}}$ for connections from hidden layer tap to hidden
target.

Rectified Linear Units (ReLUs) were chosen as the activation functions of
neurons since they have less likelihood of vanishing
gradient \parencite{Nair2010}. ReLU activation function is given by:

\begin{equation}
  f(x) = max(x, 0)
\end{equation}

The output layers (both final and hidden branch) used softmax logistic
regression while the cost function was log multinomial loss. For hidden output
cost function, L2 regularization was also added for weights of hidden layer 1.
The training was performed using simple stochastic gradient descent using the
algorithm explained in Section~\ref{sec:algo} with mini batch size of 256 and
momentum value of 0.9. Since, the aim is comparison, no attempts were made to
achieve higher than the state-of-the-art accuracies.

The network was implemented using the Python library for Deep Neural Networks,
kayak \footnote{Harvard Intelligent Probabilistic Systems (HIPS),
  \url{https://github.com/HIPS/Kayak}}.

\subsection{Performance}

Three training experiments were performed, as elaborated below:

\begin{enumerate}[1.]
\item \emph{\textbf{With} simultaneous updates for the shared layers (100
    epochs) + fine tuning (20 epochs)}
\item \emph{\textbf{Without} simultaneous updates for shared layer by ignoring
    gradients coming from hidden layer target (100 epochs) + fine tuning (20
    epochs)}
\item \emph{Training only using the hidden layer target (100 epochs) + fine
    tuning (20 epochs)}
\end{enumerate}

The fine tuning step only updates the hidden tap to hidden target weights and
biases, $\boldsymbol{W_{H}, b_{H}}$. This was performed to see the state of the
losses of the network with respect to the hidden layer targets. All the three
training experiments were performed with the same set of hyper-parameters and
were repeated 20 times to account for the random variations. Values of mean
training losses throughout the course of training were plotted using all 20
repetitions.

\begin{figure*}[h]
  \centering
  \includegraphics[width=\columnwidth]{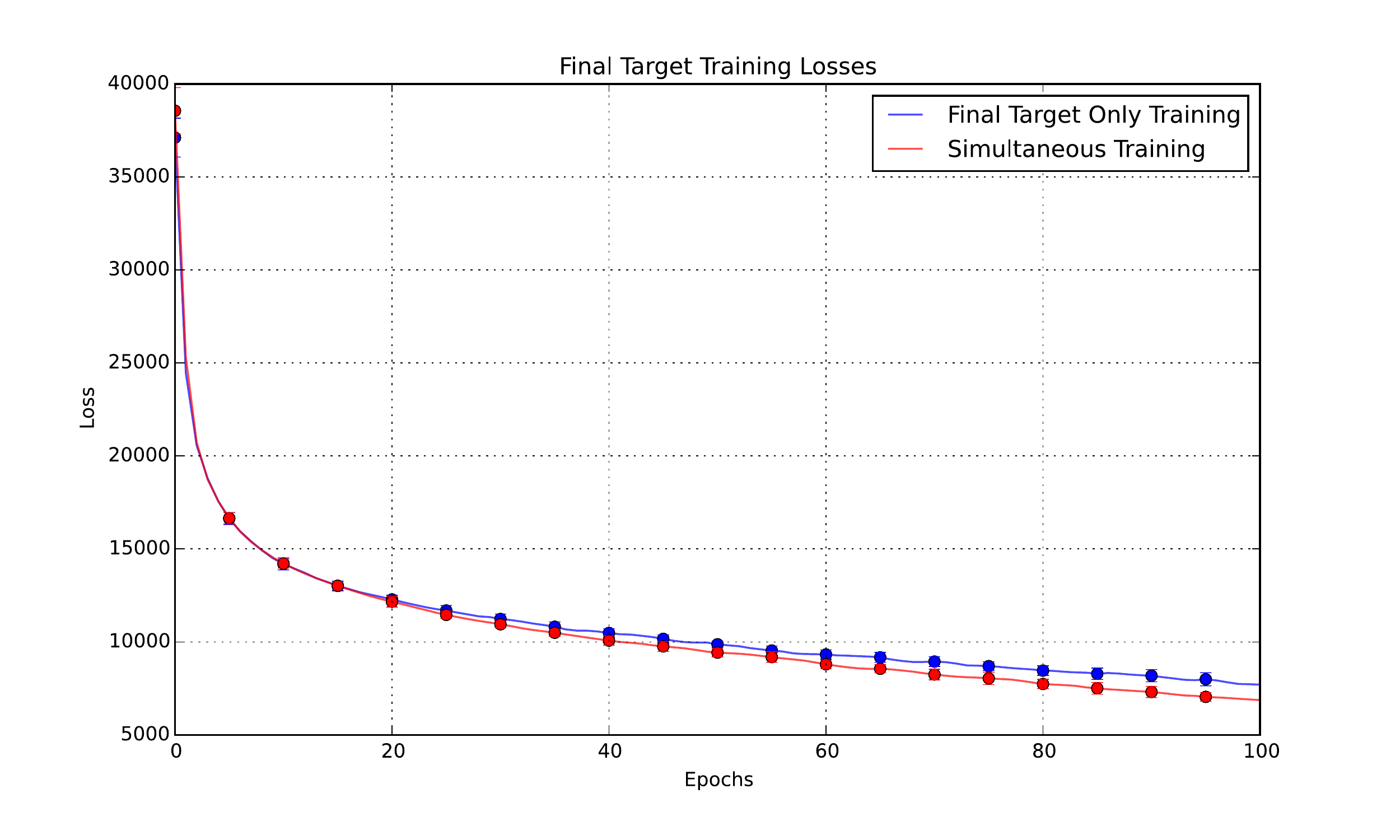}
  \caption{Mean final target losses during training. Errorbars represent one
    standard deviation.}
  \label{fig:targetloss}
\end{figure*}

The plot of training losses for final layer target in experiment 1 and 2 is
shown in Figure~\ref{fig:targetloss}. From the plot, simultaneous training is
seemingly performing better than direct training involving only target cost
function minimization.

\begin{figure*}[h!]
  \centering
  \includegraphics[width=\columnwidth]{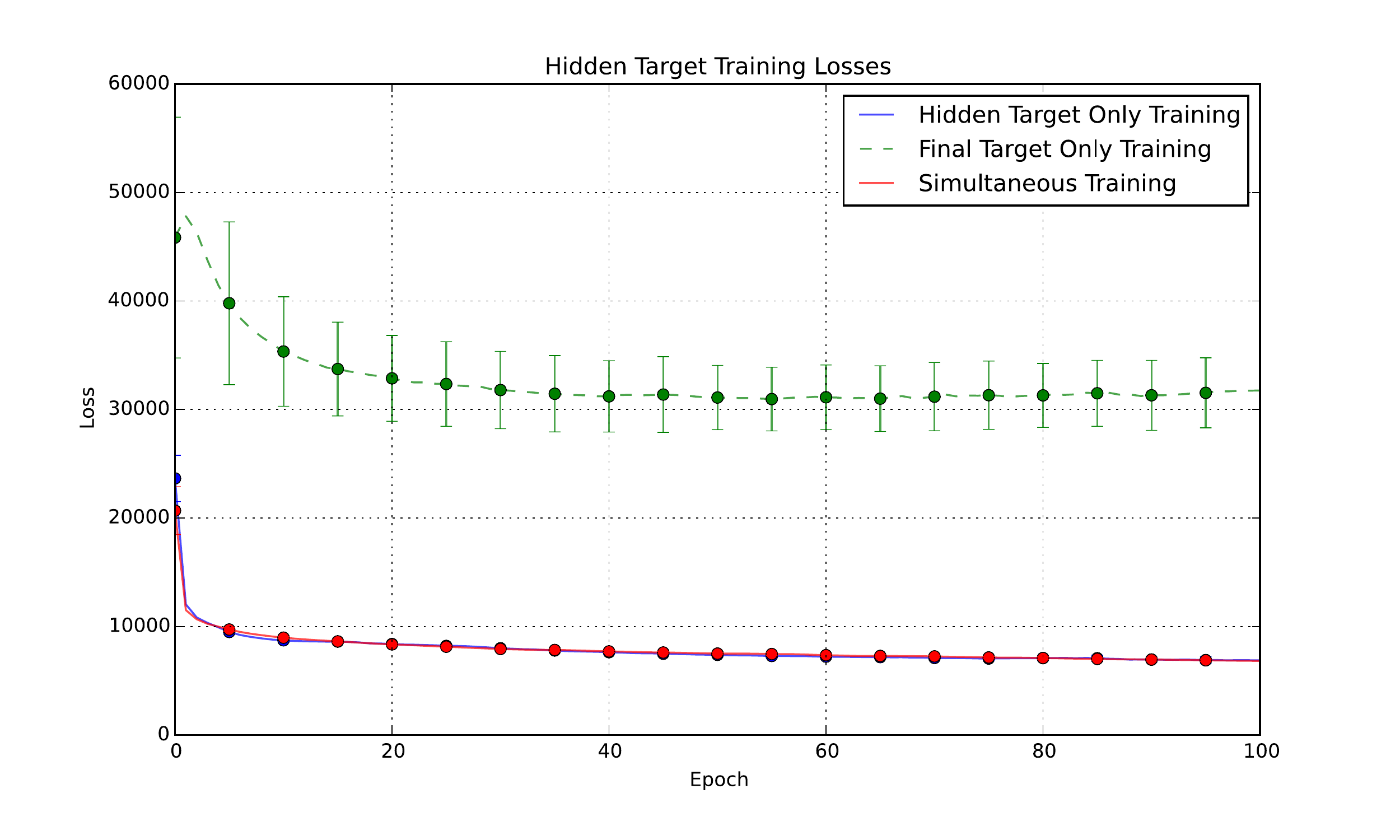}
  \caption{Mean hidden target losses during training. Errorbars represent one
    standard deviation.} 
  \label{fig:hiddenloss}
\end{figure*}

\begin{figure*}[h!]
  \centering
  \includegraphics[width=\columnwidth]{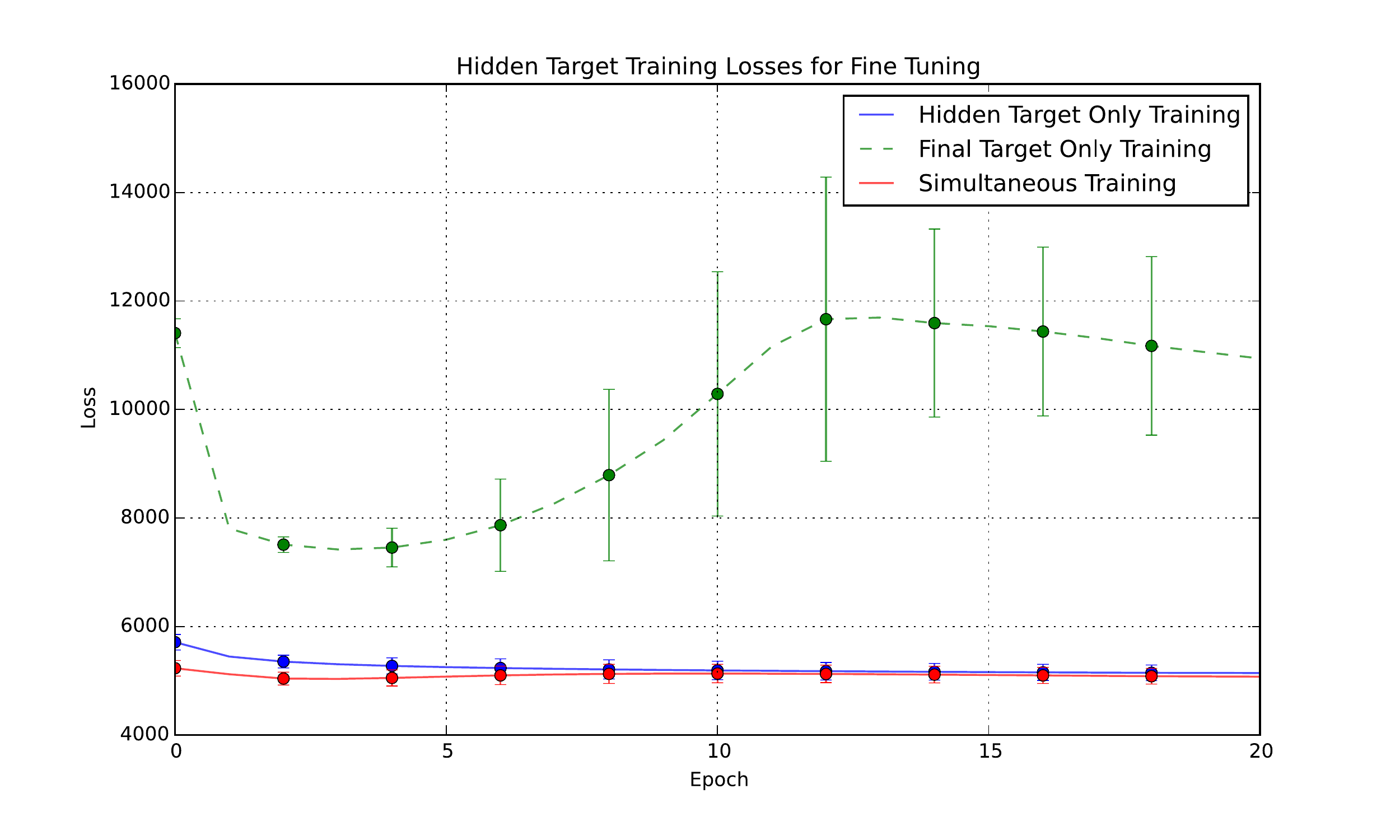}
  \caption{Mean hidden target losses during fine tuning. Errorbars represent one
    standard deviation.} 
  \label{fig:hiddenlosstune}
\end{figure*}

Plot of training losses for hidden layer target in all three experiments is
given in Figure~\ref{fig:hiddenloss}. Here, training with only minimization of
final cost is not able to generate enough effective representation of data to
help in minimization of hidden cost function, while simultaneous training and
training involving only hidden cost minimization are giving almost similar
performance. The situation is clearer in Figure~\ref{fig:hiddenlosstune}, which
is plot of losses for hidden target during the fine tuning process for all the
three experiments. As this graph shows, training only with final target cost in
consideration is not able minimize loss well as compared to other two methods.
Also, curve of simultaneous training starts with lesser loss than curve of
training with hidden cost only. This depicts better updates of weights in
simultaneous training as compared to training with only hidden cost.

\begin{figure*}[h!]
  \centering
  \includegraphics[width=\columnwidth]{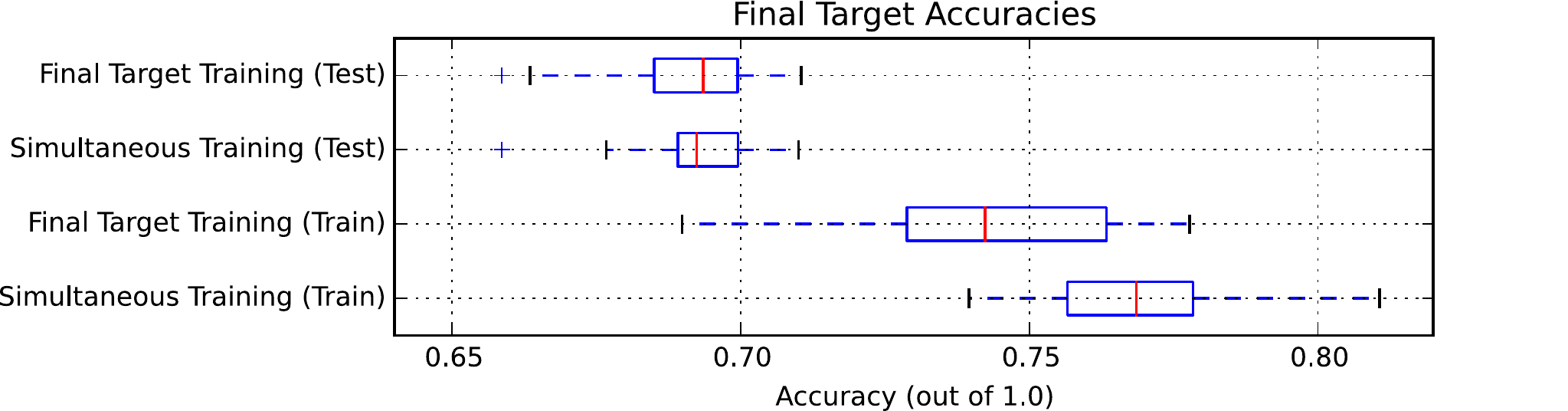}
  \caption{Boxplots for final target accuracies}
  \label{fig:boxtarget}
\end{figure*}

\begin{figure*}[h!]
  \centering
  \includegraphics[width=\columnwidth]{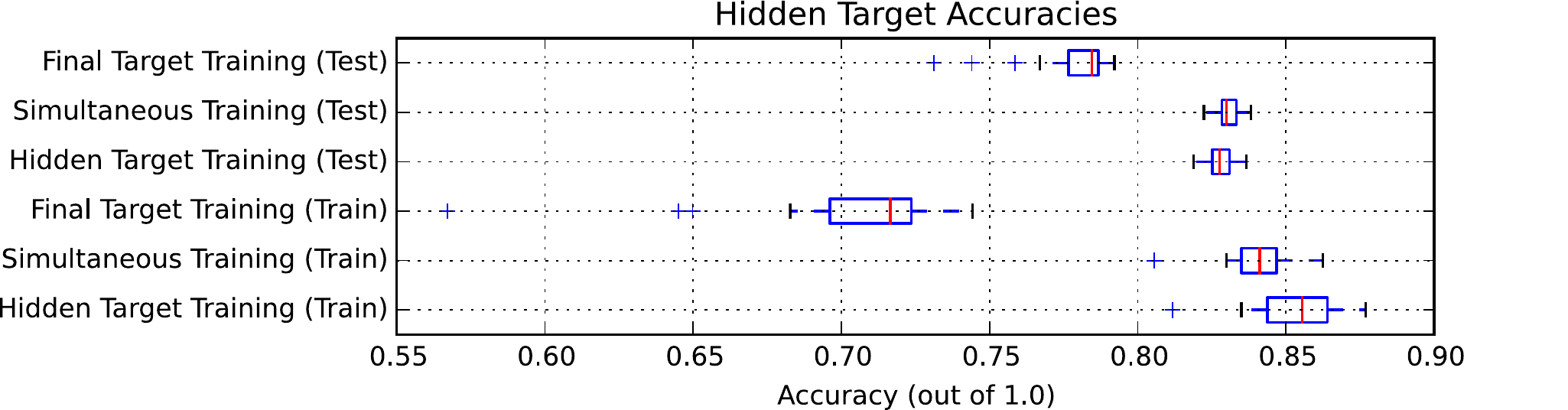}
  \caption{Boxplots for hidden target accuracies}
  \label{fig:boxhidden}
\end{figure*}

Figure~\ref{fig:boxtarget} and~\ref{fig:boxhidden} show box plots of the
accuracies over the 20 repeated experiments for hidden and final targets.

Table~\ref{tbl:accuracies} shows the mean classification accuracy on final and
hidden target for both training and testing set. As clear from the table and box
plots, the simultaneous training is providing better performance than other
training methods.

\begin{table}[h!]
\centering
\resizebox{\textwidth}{!}{%
\begin{tabular}{lll}
 & \multicolumn{2}{c}{\textbf{Hidden Target Accuracy}} \\
 & \multicolumn{1}{c}{\textbf{Train (\%)}} & \multicolumn{1}{c}{\textbf{Test (\%)}} \\ \hline
\textbf{Hidden Training} & \textbf{85.320316 (1.523254)} & 82.822154 (0.417403) \\
\textbf{Final Training} & 70.205044 (4.088195) & 77.763680 (1.602464) \\
\textbf{Simultaneous Training} & 84.051977 (1.182006) & \textbf{83.052736 (0.356259)} \\
 &  &  \\
 & \multicolumn{2}{c}{\textbf{Final Target Accuracy}} \\
 & \multicolumn{1}{c}{\textbf{Train (\%)}} & \multicolumn{1}{c}{\textbf{Test (\%)}} \\ \hline
\textbf{Hidden Training} & 4.998253 (1.453776) & 5.015446 (1.446461) \\
\textbf{Final Training} & 74.332472 (2.295639) & 69.088703 (1.325522) \\
\textbf{Simultaneous Training} & \textbf{76.824787 (1.792208)} & \textbf{69.205649 (1.183573)}
\end{tabular}
}
  \caption{Mean accuracies for the experiments. The values in parentheses are
    standard deviations.}
  \label{tbl:accuracies}
\end{table}

\section{Conclusion}
\label{sec:conclusion}

This paper presented a branching architecture for neural networks that, when
applied to appropriate problem with multiple level of outputs, inherently cause
the hidden layers to store meaningful representations and helps in improving
performance. The training curves showed that during simultaneous training, the
shared layers were learning a representation that minimized both cost functions
as well as had better weights for hidden targets.

The branches helps in enforcing information in hidden layers and thus the
auxiliary branches can be added or removed easily from the network, this
provides flexibility in terms of modularity and scalability of network.

\section{Future Work}
\label{sec:future}

This key concept in the proposed architecture is to exploit the hidden layers by
meaningful representations. Using a hierarchy of target, the proposed
architecture can form meaningful hidden representations.

An extended experiment can be done with many branches. Convolutional networks
working on computer vision problems are ideal candidates for these tests, as it
is easy to visualize the weights to find connections with the desired
representations. Also, vision problems can be broken in many level of details
and thus a hierarchy of outputs can be generated from single output layer.

Whereas this paper focused on a problem involving branches from the hidden
layers, an exploration can be done in which few hidden neurons directly
represent the hidden targets without any branching. Further, work can be done
for construction of multiple level of outputs from single output. This can be
useful for computer vision problems, where different level of outputs can be
practically useful.

\printbibliography{}

\end{document}